# VTFusion: A Vision-Text Multimodal Fusion Network for Few-Shot Anomaly Detection

Yuxin Jiang, Yunkang Cao, Yuqi Cheng, Yiheng Zhang, Weiming Shen, *Fellow, IEEE*

*Abstract*—Few-Shot Anomaly Detection (FSAD) has emerged as a critical paradigm for identifying irregularities using scarce normal references. While recent methods have integrated textual semantics to complement visual data, they predominantly rely on features pre-trained on natural scenes, thereby neglecting the granular, domain-specific semantics essential for industrial inspection. Furthermore, prevalent fusion strategies often resort to superficial concatenation, failing to address the inherent semantic misalignment between visual and textual modalities, which compromises robustness against cross-modal interference. To bridge these gaps, this study proposes VTFusion, a vision-text multimodal fusion framework tailored for FSAD. The framework rests on two core designs. First, adaptive feature extractors for both image and text modalities are introduced to learn task-specific representations, bridging the domain gap between pre-trained models and industrial data; this is further augmented by generating diverse synthetic anomalies to enhance feature discriminability. Second, a dedicated multimodal prediction fusion module is developed, comprising a fusion block that facilitates rich cross-modal information exchange and a segmentation network that generates refined pixel-level anomaly maps under multimodal guidance. VTFusion significantly advances FSAD performance, achieving image-level AUROCs of 96.8% and 86.2% in the 2-shot scenario on the MVTec AD and VisA datasets, respectively. Furthermore, VTFusion achieves an AUPRO of 93.5% on a real-world dataset of industrial automotive plastic parts introduced in this paper, further demonstrating its practical applicability in demanding industrial scenarios.

## I. Introduction

Anomaly detection (AD) aims to identify unusual instances with significant deviations from the majority of the data [1]. In industrial settings, although AD has been demonstrated to be valuable in enhancing product quality control [2]-[4], industrial AD remains challenging due to the following two reasons. Firstly, abnormal data are naturally rare, making it arduous to obtain a substantial amount of such data [5], [6]. Secondly, the collection of comprehensive and high-quality normal images for each product category under inspection requires extensive resources and time [7]. To address these issues, few-shot anomaly detection (FSAD) exclusively relying on a small amount of normal data to develop AD models has gained increasing popularity.

In the field of FSAD, two distinct frameworks have been proposed, including a vision-based framework and a multimodal-based framework. As depicted in Fig. 1 (a), the vision-based framework depends on visual information extracted from RGB images, such as texture and semantics, to generate normal prototypes, which are then used as templates for AD [8]. However, due to the unavailability of abnormal data during the training process, this framework often fails to provide a comprehensive understanding of the inherent characteristics of anomalies, leading to implicit decision boundaries between normal and abnormal images [9].

To obtain more general and commonsense information about anomalies, recent studies have paid increasing attention to a multimodal-based framework [10]. This framework aims to leverage open-vocabulary knowledge inherited from vision-language models (VLMs) such as CLIP [11] to improve the vision-based framework [12]. As shown in Fig. 1 (b), the multimodal-based framework consists of a textual part and a visual part. On the one hand, it classifies images as either normal or abnormal based on the correlations between the input images and sets of textual descriptions that delineate normal and abnormal traits, resulting in text-guided predictions [13]. On the other hand, it employs visual feature alignment to identify irregular and rare patterns, thereby generating vision-based predictions. The text-guided and vision-based predictions are then combined to improve the anomaly detection results. Nevertheless, a domain gap often exists between the natural images used to pre-train foundational models and the target FSAD domain, resulting in extracted features that may lack sufficient discriminability for distinguishing normal from anomalous samples. In addition, direct concatenation of predictions from diverse modalities may disturb the multimodal information, resulting in degraded anomaly detection performance [14].

Here, we propose VTFusion, a vision-text multimodal fusion framework (see Fig. 1 (c)), which leverages complementary information from visual and text modalities to detect anomalies. The framework consists of two main stages: (i) extraction of adaptive features from visual and textual inputs to produce

This work was supported by the Fundamental Research Funds for the Central Universities of China under Grant HUST:2021GCRC058 *(Corresponding author: Weiming Shen)* and was part by the HPC Platform of Huazhong University of Science and Technology where the computation is completed.
The authors are with the National Center of Technology Innovation for Intelligent Design and Numerical Control，Huazhong University of Science and Technology, Wuhan 430074, China (e-mail: yuxinjiang@hust.edu.cn; cyk_hust@hust.edu.cn; yuqicheng@hust.edu.cn; m202270733@hust.edu.cn; wshen@ieee.org).

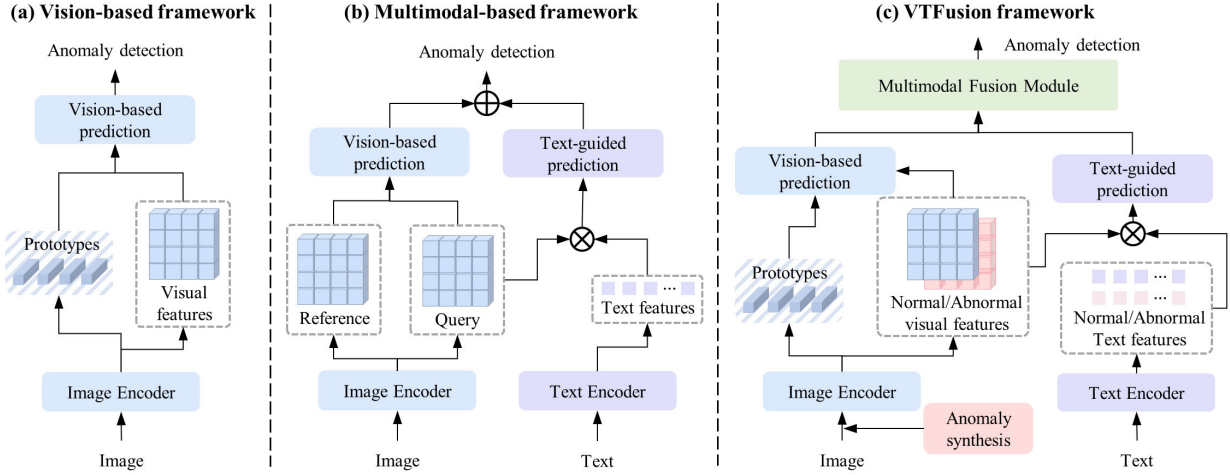

**Fig. 1.** Comparisons of different FSAD frameworks.

single-modal prediction results. (ii) effective cross-modal fusion to generate final multimodal predictions.

Specifically, to bridge the domain gap between pretrained datasets and the target industrial data, the framework employs adaptive CLIP images and text encoders during feature extraction. These encoders facilitate the transfer of dominant features from the source domain to the target industrial domain, thereby generating features enriched with domain-specific information. Furthermore, this study synthesizes diverse anomalies to establish more compact normal feature distributions and then obtain more discriminative features. The improved descriptiveness of both visual and textual features leads to superior single-modal prediction results.

At the multimodal fusion stage, different from previous methods that directly concatenate single-modal predictions, VTFusion introduces a dedicated multimodal prediction fusion module. This module enables semantic alignment between visual and textual features while facilitating information exchange between commonsense knowledge encoded in the text modality and domain-specific visual cues. Additionally, a segmentation network is proposed, which is trained with pixel-level supervision to produce precise anomaly maps.

The contributions of this paper can be summarized as follows:
- This study proposes VTFusion, a vision-text multimodal FSAD method that combines commonsensical knowledge and domain-specific information by fusing vision-based and text-guided prediction maps, enabling a comprehensive understanding of the anomaly patterns.
- This study develops an adaptive CLIP to mitigate the domain gap. Moreover, by the incorporation of four distinct types of synthetic anomalies, the adaptive CLIP is further enhanced for explicitly establishing separating boundaries.
- This study introduces a multimodal prediction fusion module to mitigate cross-modal interference. Integrating dedicated fusion blocks and a segmentation network, the proposed VTFusion framework learns discriminative hybrid features and generates fine-grained anomaly detection results.

The following sections of this paper are organized as follows: Section II offers an in-depth exploration of current literature in the field. Section III provides a detailed explanation of the proposed VTFusion method. Section IV delves into the presentation and analysis of comprehensive evaluation results derived from rigorous assessments on two publicly available datasets and a real-world application involving automotive plastic parts. Finally, Section V concludes the paper and discusses the potential directions for future research.

## II. RELATED WORK

### A. Anomaly Detection

The scarcity of abnormal data typically renders anomaly detection an unsupervised learning problem [15]-[17]. These methods train exclusively on normal samples to model category-specific normality, thereby enabling the identification of anomalies [18]. Mainstream AD methods are commonly categorized into three types: feature embedding-based methods, reconstruction-based methods, and knowledge distillation-based methods [19]. Feature embedding-based methods typically rely on models pretrained on large-scale natural image datasets such as ImageNet [20] to obtain normal features. Then, statistical techniques such as normalizing flows [21], [22], multivariate Gaussian distributions [23], and memory banks [24] are employed to estimate the distributions of these normal features. Anomalies are identified by measuring deviations from the learned distribution. Reconstruction-based and knowledge distillation-based methods, on the other hand, utilize trainable networks to regress normal features. Reconstruction-based methods typically employ autoencoders [25], in which an encoder compresses normal images into latent representations and a decoder reconstructs the input [26], [27]. Knowledge distillation-based methods train a student network to mimic the intermediate representations of a pretrained teacher network [28], [29]. Since these networks are trained solely on normal data, they are expected to exhibit large regression errors for abnormal samples [30]. However, due to imprecise modeling of normality boundaries, these methods may assign unexpectedly low anomaly scores to certain anomalies [31], [32]. To address this challenge, both MemKD [33] and MRKD [19] introduce synthetic anomalies to optimize the student networks, thereby enhancing the normality of features. Specifically, these methods align the normal regions of the features with the input image, while restoring abnormal regions to a more normal appearance. Therefore, the discrepancy between the original and restored images can be effectively exploited for anomaly localization. This study extends the anomaly synthesis method to multimodal settings and achieves impressive results.

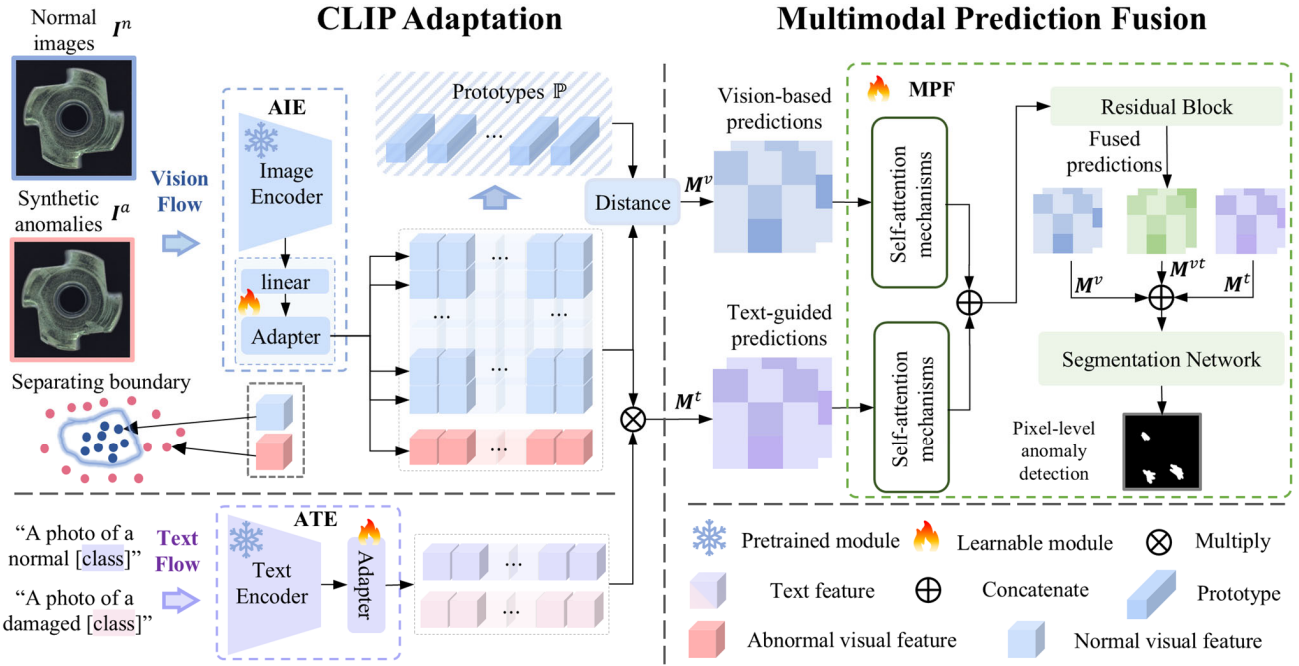

**Fig. 2.** The diagram of the proposed VTFusion model. Given a set of normal images, multiple artificial abnormal images are synthesized, supporting the AIE module in generating representative prototypes and contributing to the creation of visual-based predictions. Additionally, defined prompts are paired with images to produce text-guided predictions. Both sets of prediction maps are then sent to the MPF module for fusion and segmentation, leading to fine-grained anomaly localization results.

*B. Few-shot Anomaly Detection*

FSAD aims to train models with a limited number of normal samples [34]. However, insufficient normal data hinders accurate modeling of the full normal distribution, impeding the construction of discriminative normality templates [8]. Recent advances have mitigated this challenge. RegAD [35] employs registration-based proxy tasks to learn descriptive features by aligning intra-category samples via geometric transformations. GraphCore [8] employs a vision isometric invariant graph neural network to extract rotation-invariant structural features as the anomaly measurement features. Additionally, FastRecon [36] utilizes a regression algorithm with distribution regularization to estimate optimal transformations from support to query features. This process facilitates the reconstruction of query features in their normal versions, allowing anomalies to be identified through image alignment. Furthermore, WinCLIP [12] pioneers the application of VLMs in FSAD by utilizing CLIP to measure patch-level similarity between image and textual captions, as well as with a few reference normal images. To enhance generalization to anomalies across various applications, InCTRL [37] detects residuals between test and support images, then evaluates their similarity to textual descriptions for anomaly scoring. Despite strong performance, CLIP-based methods remain limited in industrial AD tasks due to the lack of domain-specific pretraining [13]. Adapting CLIP to industrial domains via dedicated adapters has thus attracted considerable attention [38]-[40]. In contrast to prior approaches that simply concatenate multimodal predictions, the proposed method introduces a novel pipeline that utilizes CLIP and a multimodal prediction fusion scheme to exploit the complementary strengths of visual and textual modalities for superior anomaly detection.

## III. PROPOSED METHOD

*A. Problem Definition and Method Overview*

**Problem Statement.** This study aims to develop a specialized model for AD, particularly focusing on scenarios characterized by limited training data. The model is trained on a small dataset denoted as $\mathcal{I}_{train}$, which comprises a sparse number of normal images $\mathcal{I}^n = \{I_i^n\}_{i=1}^{N_0}$, where $N_0 \leq 8$ in this study. After training, the model is expected to accurately score anomaly severity and precisely localize abnormal regions in unseen images from the testing dataset ($\mathcal{I}_{test}$).

**Overview of the Proposed Method.** The proposed VTFusion is a multimodal framework that fuses vision- and text-based anomaly predictions by utilizing both RGB images and corresponding text information. An overview of the method is presented in Fig. 2, which comprises two parallel flows: a vision flow and a text flow. In the vision flow, we first synthesize abnormal images $I^a$ based on the normal images $I^n$.

These two types of images are then input into an adaptive image encoder (AIE) to establish compact and explicit separating boundaries. This process yields discriminative visual features and a set of prototypes $\mathbb{P}$. These prototypes are then compared with the discriminative visual features to produce vision-based predictions $M^v$. Meanwhile, in the text flow, an adaptive text encoder (ATE) extracts semantic embeddings from predefined textual prompts. The resulting text features are aligned with the visual features to generate text-guided predictions $M^t$. To effectively fuse the vision-based and text-guided predictions, a multimodal prediction fusion module is incorporated, producing fused predictions $M^{vt}$. These fused predictions are subsequently employed in a segmentation block to obtain fine-grained anomaly localization results.

The following subsections detail the diversified anomaly synthesis module, the adaptive image and text encoder module, and the multimodal prediction fusion module.

*B. Diversified Anomaly Synthesis Module*

To enhance the model's comprehension of anomaly characteristics, this paper proposes the diversified anomaly synthesis (DAS) module that simulates four distinct anomaly types. By considering attributes such as spatial distribution, patterns, shapes, and textures, DAS generates synthetic anomalies that closely resemble real-world defects. Specifically, the DAS module synthesizes misplaced-type and blurry-type anomalies to mimic structural anomalies, while generating crack-type and noise-type anomalies to simulate textural anomalies. Examples of these synthetic anomalies are shown in Fig. 3 and described below.

**Misplaced-type Anomalies** are generated through a cut-and-paste operation, where some patches are randomly cut from the original images and then placed at different locations. This process effectively mimics structural misplacement and semantic irregularities while preserving local textures.

**Blurry-type Anomalies** simulate blurred areas within the image by applying a Gaussian blur, representing image degradation and reduced clarity in object details.

**Crack-type Anomalies** are synthesized by generating multiple crack-like masks with varied shapes, lengths, sizes, and textures consistent with the background pattern. These masks are then superimposed onto the foreground of the original normal images to replicate irregular fractures or material defects.

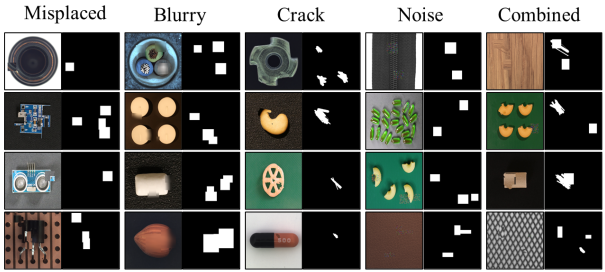

**Fig. 3.** Samples of the synthetic anomalies. The combined-type refers to samples that exhibit the presence of multiple anomaly types, including misplaced, blurry, crack and noise types.

**Noise-type Anomalies** are introduced by adding various noise distributions (*e.g.*, uniform noise, positively distributed noise, or Gaussian noise) to specific areas of normal images. This process simulates unwanted pixel intensity variations commonly caused by sensor imperfections, signal interference, or transmission errors.

*C. Adaptation of CLIP for Anomaly Detection*

To generate domain-specific and discriminative visual and text features, CLIP's feature distribution is adapted to the industrial domain. This adaptation involves the construction of adaptive image and text encoders, creating initial prototypes, and then learning the separating boundaries between normal and abnormal features using a pair of contrastive losses.

**Adaptive Image Encoder (AIE).** The image encoder of CLIP possesses a rich semantic understanding, rendering it suitable for deployment in AD. Given that the CLIP model is designed for classification and its features cannot be directly compared to text features, an additional linear layer is applied to map the image features into a joint embedding space, and then compare them with the text features [10]. Nevertheless, the domain gap between CLIP's pretraining data and target industrial data often renders features from linear layers insufficiently descriptive. To address the issue, this study introduces a feature adaptor, an additional convolutional layer placed at the end of the linear layer, to ensure the generation of more representative visual features. Specifically, given an image pair $\{I^n, I^a\}$, where $I \in \mathbb{R}^{H \times W \times 3}$, the frozen CLIP's image encoder and trainable linear layer are exploited to extract multi-level features. Subsequently, the feature adaptor facilitates the transfer of encoded features from CLIP's embedding space to a domain-specific space, yielding $\{F_l^{v_n}, F_l^{v_a}\}$, where both $F_l^{v_n}$ and $F_l^{v_a}$ are normal and abnormal features, respectively, and $l$ indicates the stage at which these features are extracted. These extracted domain-specific feature pairs play a pivotal role in enhancing VTFusion's discriminability through prototype-based comparisons, as detailed next.

**Initial Prototypes.** As illustrated in Fig. 4, we stack the normal features $\{F_1^{v_n} \cdots F_l^{v_n} \cdots F_L^{v_n}\}$, $F_l^{v_n} \in \mathbb{R}^{h \times w \times c}$ across all stages to generate $F^{v_n} \in \mathbb{R}^{h \times w \times (c \times L)}$, where $h$ and $w$ represent the height and width of the feature map, $c$ is the number of channels in each layer, and $L$ is the number of layers. In this manner, given a batch of input images, all their corresponding stacked features can be computed. Subsequently, these stacked features are averaged across the batch and compressed as $F^{\text{AVE}} \in \mathbb{R}^{N \times (c \times L)}$, where $N = h \times w$ represents the total number of locations in the feature map. The compressed feature map is utilized to initialize the prototype set $\mathbb{P}$ defined as:

$$\mathbb{P} = \bigcup_{n \in \{1, \cdots N\}} F_n^{\text{AVE}} \qquad (1)$$

Where $F_n^{\text{AVE}}$ denotes the $n$-th feature slices, and the prototypes set $\mathbb{P}$ comprises $N$ prototypes $P \in \mathbb{R}^{c \times L}$, acting as frozen anchors throughout subsequent model training. Then, we utilize the prototype set $\mathbb{P}$ to establish explicit and compact separating boundaries between normal and abnormal features.

**Separating boundaries.** The learned prototypes enable a more compact representation of the normal feature space. Specifically, given a normal feature $F^{v_n}$, we can find the closest prototype $P^*$ by calculating the L2 distance between $F_{x,y}^{v_n}$ and each of the prototypes $P_n$. We define the closest prototype as

$$P^* = \underset{P_n \in \mathbb{P}}{\operatorname{argmin}} \left\| F_{x,y}^{v_n} - P_n \right\|_2 \quad (2)$$

where $F_{x,y}^{v_n} \in \mathbb{R}^{c \times L}$ denote the $c \times L$-dimensional feature patch at positions $x \in \{1, \cdots h\}, y \in \{1, \cdots w\}$.

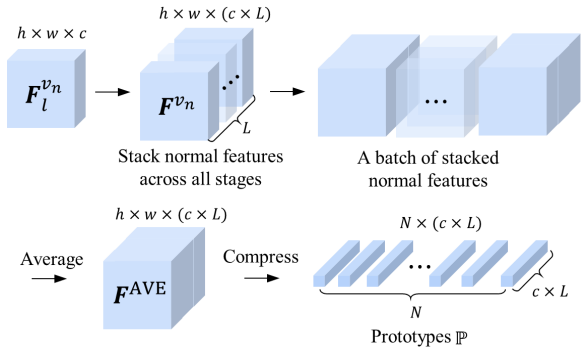

**Fig. 4.** Illustration of the initialization of prototypes.

To enclose normal features within a compact hypersphere, the closest prototypes are used as centroids of the normal feature distribution. Normal features are then encouraged to lie in close proximity to these centroids, with distances not exceeding a predefined radius $r$. This process can be achieved with a Normal Feature Compactness (NFC) loss:

$$\mathcal{L}_{NFC} = \frac{1}{N} \sum_x^h \sum_y^w \max\{0, \mathcal{D}(F_{x,y}^{v_n}, P^*) - r^2\} \quad (3)$$

where $\mathcal{D}(.,.)$ is a predefined distance metric, *i.e.*, Euclidean distance, in this paper.

However, certain anomalous features resemble normal ones and may fall near the boundary or within the tail of the normal distribution, increasing the risk of misclassification. To further refine the established boundary, an abnormal features separation (AFS) loss function is introduced to push anomalous features away from normal clusters. The AFS loss function is defined as:

$$\mathcal{L}_{AFS} = \frac{1}{N} \sum_x^h \sum_y^w \max\{0, (r + \alpha)^2 - \mathcal{D}(F_{x,y}^{v_a}, P^*)\} \quad (4)$$

where $F^{v_a} \in \mathbb{R}^{h \times w \times c \times L}$ represents a concatenated feature derived from $\{F_1^{v_a} \cdots F_l^{v_a} \cdots F_L^{v_a}\}$, and $F_{x,y}^{v_a} \in F^{v_a}$. $\alpha$ is a radius relaxation coefficient.

After obtaining a well-defined boundary that separates normal and abnormal features, the extracted feature $F^v$ of the training images are sufficiently discriminative. They can be compared with the prototypical feature $P^*$ to generate a vision-based prediction, denoted as $M^v$, which offers a quantitative measure to assess the likelihood of input images being abnormal, and is formulated as follows:

$$M^v = \mathcal{D}(F^v, P^*) \quad (5)$$

**Adaptive Text Encoder (ATE).** In addition to harnessing CLIP's image encoder, our approach further exploits the powerful generalization capability offered by natural language through CLIP's text encoder. Following [10], we create textual descriptions for both normal and abnormal images, such as "a photo of a normal [o]" or "a photo of a damaged [o]", where [o] is an object-level label, *e.g.*, "bottle". These descriptions serve as the input to the CLIP's text encoder. To bridge the domain gap, we integrate a text adapter, *i.e.*, an MLP layer, to refine the output of the CLIP's text encoder, ultimately yielding a text feature $F^t \in \mathbb{R}^{2 \times c_{text}}$. Then, a text-guided prediction is computed using the following formulation:

$$M^t = \sum_l \operatorname{softmax}(F_l^v (F^t)^T) \quad (6)$$

where $M^t$ represents the text-guided prediction.

*D. Multimodal Prediction Fusion Module*

Current multimodal methods directly concatenate vision-based and text-guided predictions to detect anomalies, which compromises robustness against cross-modal interference. To address this issue, we introduce the Multimodal Predictions Fusion (MPF) module that integrates multimodal predictions while learning their interdependencies. The MPF module consists of two main components: a fusion block and a segmentation network.

**Fusion Block.** The fusion block aims to unify the distribution of multimodal predictions. Specifically, we denote the vision-based and text-guided predictions of a pair as $\{M^v, M^t\}$. The self-attention mechanism is employed on the $\{M^v, M^t\}$ to capture the relative importance and relevance of distinct regions or features. By calculating attention matrices for each input modality independently, the self-attention mechanism highlights discriminative regions that significantly contribute to the anomaly detection task. Subsequently, a softmax function is applied to compute vision-based and text-guided attention matrices. Furthermore, to extract interaction information between the two modalities, the two attention matrices are concatenated in a channel-wise manner. The resulting combined features are then passed through a residual block, denoted as $\mathcal{G}$. This process ultimately yields the output as a fused visual-text prediction, denoted as $M^{vt}$:

$$M^{vt} = \mathcal{G}(\operatorname{SAM}(M^v) \oplus \operatorname{SAM}(M^t)) \quad (7)$$

where $\mathcal{G}$ represents the residual block that consists of convolutional operations, batch normalization, and ReLU activation. The symbol $\oplus$ denotes a channel-wise concatenation operation, and $\operatorname{SAM}(\cdot)$ refers to the self-attention mechanism.

**Segmentation Network.** To leverage the vision-text predictions for generating fine-grained pixel-level anomaly maps, this study employs an FPN-like [41] segmentation network, denoted as $\phi$. Even though the fusion block promotes the cross-modal interactions, some information loss persists during fusion. To preserve complementary details, we concatenate the vision-based prediction $M^v$, the text-guided prediction $M^t$ and the vision-text prediction $M^{vt}$ as input to the segmentation network. To effectively detect anomalies of varying scales, a multi-scale feature aggregation mechanism [42] is integrated to extract features at multiple scales from the concatenated input. Then, a loss function, denoted as $\mathcal{L}_{SEG}$, is applied to supervise the generation of anomaly maps, which indicate the likelihood of individual pixels being categorized as anomalies. The segmentation loss function is defined as follows:

$$\mathcal{L}_{SEG} = \frac{1}{H \times W} \sum_x^H \sum_y^W (\boldsymbol{M}_{x,y}^f - \boldsymbol{M}_{x,y}^g)^2 \quad (8)$$

$$\boldsymbol{M}^f = \text{Upsample}(\phi(\boldsymbol{M}^v \oplus \boldsymbol{M}^t \oplus \boldsymbol{M}^{vt})) \quad (9)$$

where $\boldsymbol{M}^g \in \mathbb{R}^{H \times W}$ is the ground truth mask of synthetic anomalies. The Upsample(·) function is utilized to interpolate the output of the segmentation network to match the resolution of the input images, resulting in anomaly score maps denoted as $\boldsymbol{M}^f \in \mathbb{R}^{H \times W}$. During the testing phase, the output $\boldsymbol{M}^f$ is utilized as pixel-wise anomaly scores, and the maximum value in $\boldsymbol{M}^f$ is employed as an image-level anomaly score.

The overall framework of VTfusion is trained with the total expressed as follows:

$$\mathcal{L}_{total} = \mathcal{L}_{AFS} + \mathcal{L}_{AFS} + \lambda \times \mathcal{L}_{SEG} \quad (10)$$

where $\lambda$ is a loss balancing hyper-parameter.

The process of the proposed method is summarized in Algorithm 1.

## IV. EXPERIMENTS AND ANALYSES

### A. Datasets and Evaluation Metrics

**Datasets.** The study employed the VTFusion on two prominent datasets, MVTec AD [43] and VisA [44]. MVTec AD comprises 5354 high-resolution images distributed across five texture classes and ten object classes. The test set includes normal and annotated abnormal images. On the other hand, the VisA dataset consists of 12 subsets, totaling 10821 images with 9621 normal and 1200 abnormal images. The anomalies in the VisA dataset cover surface defects, including scratches, cracks, and color spots, as well as structural defects such as misplacement or missing parts.

**Evaluation Metrics.** To evaluate the performance of the proposed method, this study employed two common metrics: Image-level Area Under the Receiver Operating Characteristics (AUROC) and Pixel-level AUROC. The former metric gauges the effectiveness of image-level anomaly detection, while the latter metric assesses the accuracy of pixel-level anomaly detection.

**Implementation Details.** In the experimental setup, we employed the OpenCLIP[1] implementation along with its pre-trained models, specifically utilizing the LAION-400M [45] based CLIP with ViT-B/16+. The hyperparameters $r$ and $\alpha$ in Eqs. (3) and (4) were fixed at 0.00001 and 0.1, respectively, while $\lambda$ in Eq. (10) retained its default value. Optimization was performed using the Adam optimizer with learning rates of $1 \times 10^{-3}$ for the AIE module and $1 \times 10^{-4}$ for the ATE and MPF modules. The implementation was carried out using the PyTorch V1.12.0 framework, and training took place on a Nvidia GeForce RTX 3090 Ti GPU and an Intel i9@3.00GHz CPU. B. Following these settings, VTFusion is trained using 2-shot, 4-shot, and 8-shot normal images. The experimental results are discussed in the next section.

---

| Algorithm 1: Training of the Proposed VTFusion |
|---|
| **Input**: Training dataset: $\mathcal{I}_{train}$; Adaptive Image Encoder: AIE; Adaptive Text Encoder: ATE; Multimodal Prediction Fusion: MPF; The hyperparameters: $r$, $\alpha$, and $\lambda$ |
| **Output**: The parameters $\{\Theta_{AIE}, \Theta_{ATE}, \Theta_{MPF}\}$ |
| 1: Initialize the parameters $\{\Theta_{AIE}, \Theta_{ATE}, \Theta_{MPF}\}$; |
| 2: **for** iteration number **do** |
| 3:    Sample a batch of training data; |
| 4:    Synthesize abnormal samples using the DAS module; |
| 5:    $\{\boldsymbol{F}_l^{vn}, \boldsymbol{F}_l^{va}\} \leftarrow$ Extract both normal and abnormal visual images; |
| 6:    $\mathbb{P} \leftarrow$ Initialize the prototype set using Eq. (1); |
| 7:    Calculate a pair of contrastive losses using Eq. (2) - (4); |
| 8:    $\boldsymbol{M}^v \leftarrow$ Generate vision-based predictions using Eq. (5); |
| 9:    $\boldsymbol{F}^t \leftarrow$ Extract text features; |
| 10:   $\boldsymbol{M}^t \leftarrow$ Generate text-based predictions using Eq. (6); |
| 11:   $\boldsymbol{M}^{vt} \leftarrow$ Generate fused vision-text predictions using Eq. (7); |
| 12:   Calculate the segmentation loss using Eq. (8) - (9); |
| 13:   Update VTFusion model using Eq. (10); |
| **end for** |

### B. Quantitative Results

Tables I provide a comparative analysis of image-level and pixel-level anomaly detection on the MVTec AD and VisA. The comparison methods include vision-based methods such as RD4AD [35], PatchCore [24], RegAD [35], and FastRecon [36], as well as multimodal-based methods like WinCLIP+ [12], RWDA [46], and InCTRL [37]. VTFusion consistently outperforms SOTA methods across 2-shot, 4-shot, and 8-shot settings. Specifically, the image-level AD performance of VTFusion exhibits remarkable enhancements compared to vision-based SOTA methods, with increases of 5.8%, 3.0%, and 2.1% (4.6%, 3.0%, 4.8%) on MVTec AD (and VisA), respectively. These results highlight the value of CLIP-derived commonsense knowledge as an effective complement to visual features in anomaly detection. Additionally, VTFusion outperforms its multimodal-based counterparts, achieving improvements of 2.8%, 2.5%, and 2.0% on MVTec AD and attaining the best performance on VisA. For pixel-level anomaly detection, VTFusion also attains leading results of 97.1% on MVTec AD and 97.5% on VisA, further underscoring the effectiveness of the proposed method.

### C. Qualitative Results

Fig. 5 presents qualitative comparisons between the proposed VTFusion method and existing approaches on the MVTec AD and VisA datasets. The visualizations clearly demonstrate the superior capabilities of VTFusion in delivering more reliable pixel-level anomaly detection results. Notably, in cases where significant inter-class variations exist, such as diverse rotational angles in screw and grid images, as well as irregular patterns in

---

[1] https://github.com/mlfoundations/open_clip

## TABLE I
COMPARISON OF K-SHOT IMAGE-LEVEL AND PIXEL-LEVEL ANOMALY DETECTION RESULTS ON THE MVTEC AD AND VISA DATASET: AUROC (%). **VISUAL-BASED METHODS ARE SHADED IN GRAY.** THE BEST AND SECOND-BEST RESULTS ARE RESPECTIVELY MARKED IN BOLD AND UNDERLINED.

| Method | MVTec AD | | | | | | VisA | | | | | |
| --- | --- | --- | --- | --- | --- | --- | --- | --- | --- | --- | --- | --- |
| | Image-level | | | Pixel-level | | | Image-level | | | Pixel-level | | |
| | 2-shot | 4-shot | 8-shot | 2-shot | 4-shot | 8-shot | 2-shot | 4-shot | 8-shot | 2-shot | 4-shot | 8-shot |
| RD4AD | 75.5 | 76.9 | 78.5 | 71.8 | 72.2 | 73.0 | 73.4 | 73.2 | 78.2 | 95.9 | 94.0 | 96.6 |
| PatchCore | 87.8 | 89.5 | 94.3 | 94.8 | 95.0 | 95.6 | 81.6 | 85.3 | 84.4 | 96.1 | 96.8 | <u>96.8</u> |
| RegAD | 85.7 | 88.2 | 91.2 | 94.6 | 95.7 | <u>96.7</u> | 55.7 | 57.4 | 58.9 | - | - | - |
| FastRecon | 91.0 | 94.2 | 95.2 | - | - | - | - | - | - | - | - | - |
| WinCLIP+ | <u>94.4</u> | 95.2 | - | <u>96.0</u> | 96.2 | - | 84.6 | 87.3 | - | **96.8** | 97.2 | - |
| RWDA | 94.0 | 94.5 | - | - | - | - | 85.6 | 86.6 | - | - | - | - |
| InCTRL | 94.0 | 94.5 | 95.3 | - | - | - | 85.8 | 87.7 | 88.7 | - | - | - |
| **VTFusion** | **96.8** | **97.0** | **97.3** | **96.3** | **97.0** | **97.1** | **86.2** | **88.3** | **89.2** | <u>96.6</u> | **97.2** | **97.5** |

## TABLE II
ABLATION STUDIES OF DIFFERENT MODULES ON MVTEC AD AND VISA FOR IMAGE-LEVEL AND PIXEL-LEVEL ANOMALY DETECTION WITH $k = 2, 4, 8$. THE BEST RESULT IS HIGHLIGHTED IN BOLD.

| Module | | | MVTec AD | | | | | | VisA | | | | | | |
| --- | --- | --- | --- | --- | --- | --- | --- | --- | --- | --- | --- | --- | --- | --- | --- |
| | | | Image-level | | | Pixel-level | | | Image-level | | | Pixel-level | | | |
| ATE | AIE | MPF | 2 | 4 | 8 | 2 | 4 | 8 | 2 | 4 | 8 | 2 | 4 | 8 |
| √ | | | 95.3 | 95.1 | 95.6 | 91.9 | 92.0 | 92.0 | 83.8 | 84.0 | 83.6 | 92.2 | 91.8 | 92.1 |
| √ | √ | | 95.4 | 94.8 | 95.3 | 95.3 | 96.2 | 96.4 | 82.9 | 82.9 | 83.2 | 94.8 | 95.0 | 94.5 |
| √ | √ | √ | **96.8** | **97.0** | **97.3** | **96.3** | **97.0** | **97.1** | **86.2** | **88.3** | **89.2** | **96.3** | **97.0** | **97.1** |

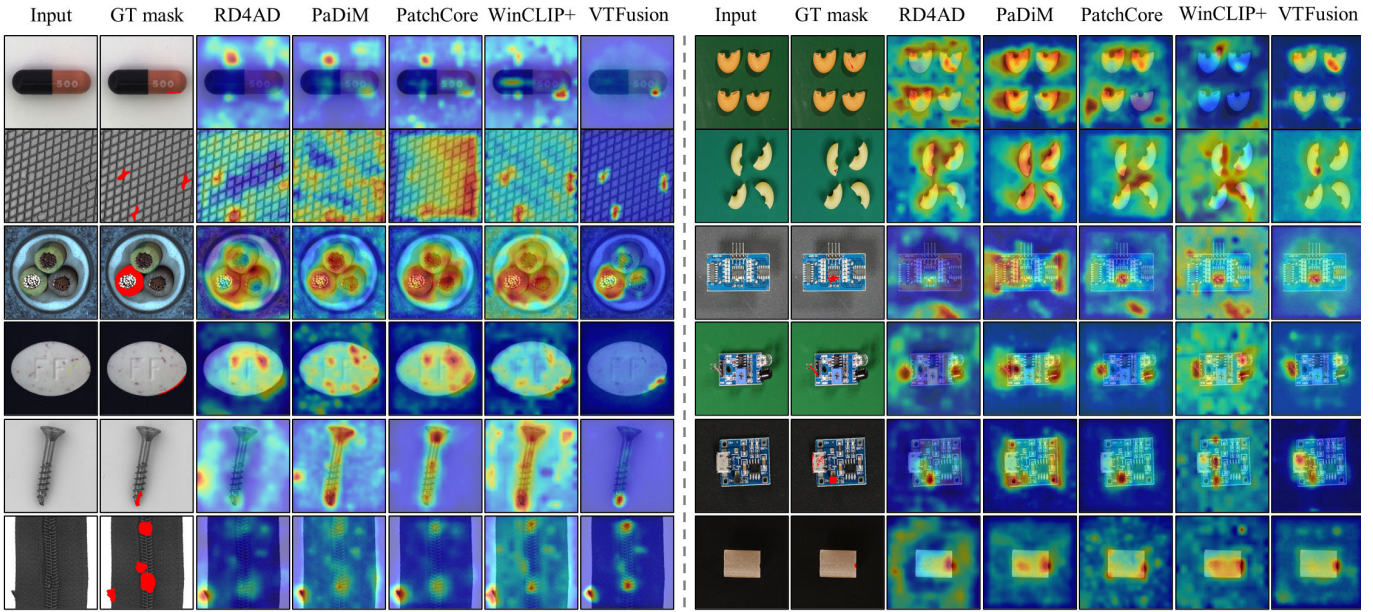

**Fig. 5.** Results visualizations of VTFusion on various categories from MVTec AD and VisA datasets. The evaluated categories, listed from up to bottom, include capsule, grid, cable, pill, screw, zipper, macaroni1, macaroni2 pcb1, pcb2, pcb4, fryum.

pill images, VTFusion diminishes the likelihood of misclassifying normal images as anomalies when compared to existing methods. This effectively mitigates the false positives, resulting in more precise pixel-level anomaly detection results. Furthermore, when confronted with disturbances like unclean backgrounds or impurities on objects (*e.g.*, pcb2 and macaroni1), VTFusion exhibits remarkable resilience, emphasizing its robustness in challenging scenarios.

*D. Ablation Study*

This study conducted extensive ablation studies on the MVTec AD and VisA datasets to quantify the contribution of each component in VTFusion. The corresponding results are presented in Table II.

TABLE III
ABLATION STUDIES OF THE DAS MODULE ON MVTEC AD AND VISA FOR IMAGE-LEVEL AND PIXEL-LEVEL ANOMALY DETECTION WITH $k = 2$.

| Types | MVTec AD | | VisA | |
|---|---|---|---|---|
| | Image | Pixel | Image | Pixel |
| Misplaced | 96.5 | 95.1 | 83.8 | 92.1 |
| Fuzzy | 96.5 | 95.3 | 85.3 | 96.5 |
| Crack | 96.3 | 95.6 | 85.8 | 95.1 |
| Noise | 96.4 | 90.4 | 85.2 | **96.6** |
| All | **96.8** | **96.3** | **86.2** | **96.6** |

**Effect of DAS.** To investigate the impact of synthesized anomalies, we compare configurations using single versus multiple anomaly types. The results, as presented in Table III, indicate that integration of multiple anomaly categories leads to superior outcomes compared to utilizing a single anomaly category alone. This underscores the significance of diverse anomaly manifestations in both image-level and pixel-level anomaly detection.

**Effect of AIE.** Table II presents the performance of the AIE module when combined with the pair of contrastive losses $\mathcal{L}_{NFC}$ and $\mathcal{L}_{AFS}$. The AIE module was specifically trained using these contrastive losses to bridge the domain gaps and establish explicit decision boundaries between normal and abnormal images. The results reveal notable enhancements in pixel-level anomaly detection. On MVTec AD, AIE brings improvements of 3.4%, 3.8%, and 3.6% in the 2-, 4-, and 8-shot settings, respectively. On VisA, gains of 2.6%, 3.2%, and 2.4% are achieved under the same conditions. To further underscore the efficacy of the adaptive CLIP's image encoder in addressing the domain gap and extracting discriminative features, Fig. 6 visualizes the features generated by the AIE module. These visualizations confirm that the CLIP's image encoder, without specific training on industrial images, yields less distinct features, posing challenges in effectively discerning between normal and abnormal patterns (Fig. 6 (b)). In contrast, the features derived from the AIE module (Fig. 6 (c)) prominently emphasize abnormal features. These findings suggest that the proposed the AIE module adeptly transfers pretrained features from the source domain to the target domain, thereby enhancing the model's capacity to distinguish abnormal features.

**Effect of MPF.** Table II demonstrates the substantial improvements brought by the MPF module in image-level anomaly detection. On MVTec AD, MPF yields gains of 1.4%, 2.2%, and 2.0% in the 2-, 4-, and 8-shot settings, respectively. On VisA, the corresponding improvements are 3.3%, 5.4%, and 6.0%. Furthermore, to illustrate the fusion capability, Fig. 7 compares pixel-level anomaly detection with and without MPF. Direct concatenation of vision-based (Fig. 7(c)) and text-guided (Fig. 7(d)) predictions (Fig. 7(e)) suffers from cross-modal interference, leading to inaccurate results due to distribution misalignment. In contrast, MPF (Fig. 7(f)) effectively aligns and integrates multimodal features in a shared space, producing cleaner and more precise anomaly maps.

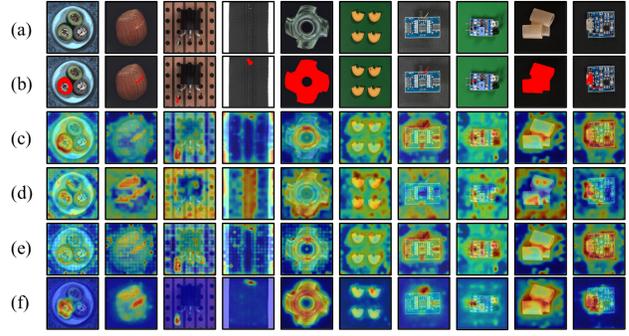

Fig. 7. Results visualizations of two types of multimodal fusions. (a) Input images. (b) ground truth masks. (c) The visualizations of visual-based predictions. (d) The visualizations of text-guided predictions. (e) The results visualizations generated by the direct concatenation of visual-based and text-guided predictions. (f) The results visualizations generated by MPF module.

*F. Anomaly Detection for Automotive Plastic Parts*

**Automotive Plastic Parts Dataset.** This section introduces the Automotive Plastic Parts Dataset (APPD), which focuses on 3D objects characterized by small, diverse, and irregularly positioned anomalies, posing a significant challenge for accurate detection [47]. To capture these complex defects, a multi-angle shooting approach was employed, involving the capture of multiple images from distinct perspectives for each component. Specifically, an inspection platform equipped with four light sources and three cameras was utilized for image acquisition (see Fig. 8 (a)). This approach resulted in the imaging of 50 plastic parts, which were subsequently segmented into 240-pixel patches after background removal.

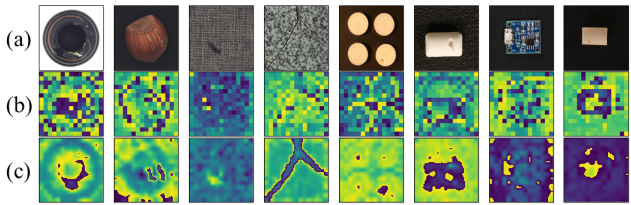

Fig. 6. The visualizations of extracted features. (a) Input images. (b) Extracted features obtained from the CLIP's image encoder with linear layers. (c) Extracted features acquired by AIE module.

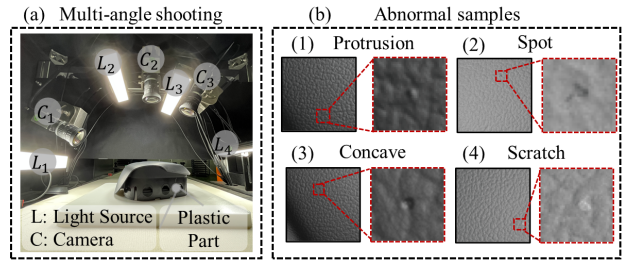

Fig. 8. (a) Schematic diagram of multi-view photography; (b) Abnormal images comprising four distinct anomaly

TABLE IV
COMPARISON OF K-SHOT ANOMALY DETECTION RESULTS ON THE APPD DATASET: PRO (%).

| Method | 2 | 4 | 8 | 10 |
|---|---|---|---|---|
| RD4AD | 67.4 | 68.4 | 77.1 | 74.7 |
| CFA | 54.2 | 54.9 | 62.3 | 84.3 |
| PatchCore | 56.2 | 62.9 | 61.6 | 75.5 |
| VTFusion | **84.5** | **89.8** | **91.6** | **93.5** |

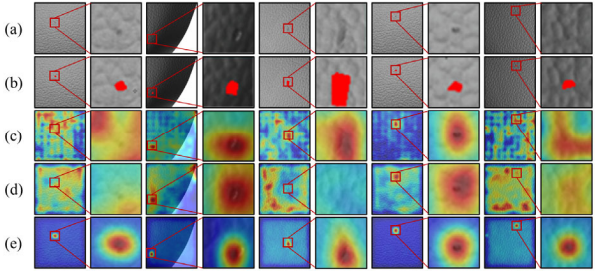

**Fig. 9.** The qualitative results of anomaly localization for VTFusion on APPD. (a) Input images. (b) Input images with ground-truth localization regions masked in red. (c) The localization results generated by RD4AD. (d) The localization results produced by PatchCore. (e) The localization results yielded by VTFusion.

This yields 4082 normal images for training, along with 1749 normal and 69 abnormal images for testing. The abnormal images comprising four different categories from the dataset are illustrated in Fig. 8 (b).

**Anomaly detection results.** The VTFusion and some comparison methods, including RD4AD, CFA [48], and PatchCore, are evaluated on the APPD. As shown in Table IV, VTFusion consistently outperforms these baselines. Specifically, in the 2-shot scenario, VTFusion achieves substantial gains of 17.1% and 28.3% over RD4AD and PatchCore, respectively. Furthermore, its performance scales effectively with increasing training samples, reaching 93.5% in the 10-shot setting. Qualitative results in Fig. 9 further demonstrate VTFusion's superiority in pixel-level anomaly detection, exhibiting precise identification of abnormal regions with minimal false positives. Overall, these results establish VTFusion as a robust solution for challenging scenarios involving subtle anomalies and high intra-class variability in limited normal samples.

*G. Limitations*

Although the proposed VTFusion has showcased remarkable performance in FSAD, there are still several areas for improvement. Firstly, VTFusion depends on manually crafted prompts for the text encoder. The process of prompt engineering requires human expertise and necessitates setting different prompts for different classes of anomalies, which conflicts with the automation requirements of industrial applications. Therefore, our objective is to improve VTFusion to distinguish between normal and abnormal samples even under more generic prompts for the text encoder (e.g., "abnormal"). Secondly, VTFusion may struggle with understanding some intricate elements, such as the macaroni and the PCB categories (as shown in Fig. 5), which may not be present in the pre-trained dataset. Addressing this challenge may entail exploring more robust foundational models.

V. CONCLUSION AND FUTURE WORK

In summary, this paper introduces VTFusion, a vision-text multimodal fusion framework designed to integrate domain-specific visual knowledge with commonsense textual semantics for enhanced few-shot anomaly detection. This method employs four targeted anomaly synthesis strategies to enable the adaptive image encoder to acquire robust domain-specific representations, while an adaptive text encoder extracts rich commonsense priors from textual descriptions. To facilitate effective multimodal integration and suppress cross-modal interference, a dedicated multimodal prediction fusion module leverages self-attention mechanisms to selectively amplify discriminative features. Collectively, these components enable VTFusion to develop a comprehensive understanding of anomalies and generate fine-grained pixel-level anomaly detection results. Experimental results demonstrate the superior performance of VTFusion, which outperforms SOTA on both MVTec AD and VisA datasets. Additionally, VTFusion is evaluated on a dataset of industrial automotive plastic parts, further confirming the practicality of this method.

Future work will focus on developing generalized multimodal fusion frameworks that combine attention mechanisms—such as cross-attention and channel attention—to effectively align and interact features across 2D, 3D, and textual modalities. Additionally, the integration of advanced anomaly synthesis methods offers substantial potential to alleviate data scarcity and further enhance the generalization capabilities of anomaly detection systems.